\documentclass[10pt,twocolumn,letterpaper]{article}

\usepackage{iccv}
\usepackage{times}
\usepackage{epsfig}
\usepackage{graphicx}
\usepackage{amsmath}
\usepackage{amssymb}
\DeclareMathOperator*{\argmax}{arg\,max}

\usepackage{multirow}
\usepackage{booktabs}
\usepackage[pagebackref=true,breaklinks=true,letterpaper=true,colorlinks,bookmarks=false]{hyperref}

\iccvfinalcopy 

\def\iccvPaperID{2315} 

\ificcvfinal\fi

\begin{document}

\title{AdaSGN: Adapting Joint Number and Model Size for Efficient Skeleton-Based Action Recognition}

\author{
	Lei Shi$^{1,2}$ \and Yifan Zhang$^{1,2}$\thanks{Corresponding Author} \and Jian Cheng$^{1,2,3}$ \and Hanqing Lu$^{1,2}$ \and
	$^1$National Laboratory of Pattern Recognition, Institute of Automation, Chinese Academy of Sciences\\
	$^2$University of Chinese Academy of Sciences\\
	$^3$CAS Center for Excellence in Brain Science and Intelligence Technology\\
	{\tt\small \{lei.shi, yfzhang, jcheng, luhq\}@nlpr.ia.ac.cn} 
}

\maketitle
\ificcvfinal\thispagestyle{empty}\fi

\begin{abstract}
Existing methods for skeleton-based action recognition mainly focus on improving the recognition accuracy, whereas the efficiency of the model is rarely considered. 
Recently, there are some works trying to speed up the skeleton modeling by designing light-weight modules. 
However, in addition to the model size, the amount of the data involved in the calculation is also an important factor for the running speed, especially for the skeleton data where most of the joints are redundant or non-informative to identify a specific skeleton.
Besides, previous works usually employ one fix-sized model for all the samples regardless of the difficulty of recognition, which wastes computations for easy samples.
To address these limitations, a novel approach, called AdaSGN, is proposed in this paper, which can reduce the computational cost of the inference process by adaptively controlling the input number of the joints of the skeleton on-the-fly. 
Moreover, it can also adaptively select the optimal model size for each sample to achieve a better trade-off between the accuracy and the efficiency. 
We conduct extensive experiments on three challenging datasets, namely, NTU-60, NTU-120 and SHREC, to verify the superiority of the proposed approach, where AdaSGN achieves comparable or even higher performance with much lower GFLOPs compared with the baseline method. 
\end{abstract}

\section{Introduction}
Action recognition is a popular research topic due to its wide range of application scenarios such as human-computer interaction and video surveillance~\cite{carreira_quo_2017,shi_gesture_2019,feichtenhofer_slowfast_2019,shi_action_2019}.
Recently, different with conventional approaches that use RGB sequences for input, exploiting the skeleton data for action recognition has drawn increasingly more attention~\cite{yan_spatial_2018,cho_self-attention_2020,shi_two-stream_2019,liu_disentangling_2020}. 
Biologically, human is able to recognize the action category by observing only the motion of joints even without the appearance information~\cite{johansson_visual_1973}.
Skeleton is such a data modality that consists of only the 2D/3D positions of the human key joints. 
Compared with the RGB data, it has smaller amount of data, higher-level semantic information and stronger robustness for complicated environment. 

\begin{figure}
    \centering
    \includegraphics[width=\linewidth]{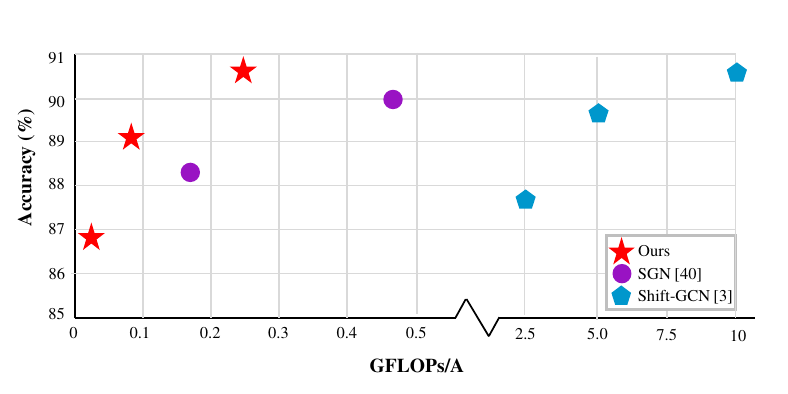}
    \caption{GFLOPs v.s. accuracy on NTU-60 (CS) dataset for state-of-the-art methods. }
    \label{fig:scatter}
\end{figure}

In early stage, approaches for skeleton-based action recognition mainly focus on designing various hand-crafted features~\cite{vemulapalli_human_2014}. 
With the success of the deep learning, various deep networks have been proposed for skeleton-based action recognition~\cite{zhang_view_2017,li_independently_2018,li_skeleton-based_2017,cao_skeleton-based_2018,yan_spatial_2018,shi_two-stream_2019}. 
However, the computational complexity of the state-of-the-art deep networks are too heavier, most of which are higher than 15 GFLOPs~\cite{cheng_skeleton-based_2020}. 
For example, the representative GCN-based work, i.e., ST-GCN~\cite{yan_spatial_2018}, costs 16.2 GFLOPs for one action sample. 
In the practical application scenario, speed is an important factor. Thus, how to speed up the current methods is a valuable-researched topic. 

Recently, some works propose to speed up the skeleton modeling by designing light-weight models, 
such as Shift-GCN~\cite{cheng_skeleton-based_2020} and SGN~\cite{zhang_semantics-guided_2020}. 
However, in addition to the model size, the amount of the data involved in the calculation is also an important factor affecting the running time, which is rarely considered in previous works. 
Especially for the skeleton data, to identify a specific action, many of the joints in the skeleton sequence are actually redundant. 
For example, to recognize the action ``swipe right", the global motion information is the key factor, which means the trajectory of the central point is discriminative enough for recognition. 
Thus, for various samples, it is unnecessary to always use all the joints for computation. 
Besides, for a specific action sequence, there are multiple stages, such as the starting stage, the course stage and the ending stage~\cite{zhao_temporal_2017}. 
Some of the stages are always non-informative, e.g., the beginning and the ending, which should not been exhaustively analysed with all of the joints. 
Furthermore, the difficulty of recognition varies for different actions. Previous works employ one fix-sized model for all the samples, which wastes computations for easy samples. 
For example, it is obvious that distinguishing the ``walking" vs. ``lying" is easier than distinguishing the ``brushing hair" vs. ``brushing teeth", so instead of using the same model, 
it is better to apply big models for hard samples whereas applying small models for easy samples. 

To address these limitations, we propose a novel approach, called adaptive SGN (AdaSGN), for efficient skeleton-based action recognition. 
SGN~\cite{zhang_semantics-guided_2020} is already a very light-weight (only 0.7M) model for skeleton-based action recognition and is the baseline method in this paper. 
Compared with SGN, our AdaSGN can further reduce more than half of the GFLOPs with even higher performance. 
In detail, AdaSGN learns a policy network to adaptively select the optimal joint number and the optimal model size to control the trade-off between the accuracy and the efficiency. 
The policy network is designed as a light-weight module and is calculated with the features of the smallest number of joints, which costs nearly no additional computational cost. 
It outputs policies according to the input sample to decide on-the-fly which model size and which joint number should be used for each skeleton. 
Because the decisions are in a discrete distribution and the decision process is non-differentiable, we employ the Straight-Through (ST) Gumbel Estimator~\cite{jang_categorical_2017} to back-propagate gradients.
Thus, the proposed method is a fully differentiable framework and can be trained in an end-to-end manner. 
The training loss is the combination of an efficiency loss and an accuracy loss, where the proportions of the two terms can be adjusted to control the accuracy-efficiency trade-off. 

To verify the superiority of the proposed approaches, extensive experiments are performed on three popular datasets, namely, NTU-60, NTU-120 and SHREC, where our method achieves comparable or even higher performance with much lower GFLOPs.
For example, we achieve +0.5\%/+0.4\% improvements of accuracy using only 39.6\%/31.3\% GFLOPs on CS/CV benchmarks of the NTU-60 dataset compared with the baseline method. 
Figure~\ref{fig:scatter} shows the GFLOPs vs. accuracy diagram on the CS benchmark of the NTU-60 dataset. 

Our contributions can be summarized as follows: 
\begin{enumerate}
    \item We propose a novel approach that can adaptively select both the optimal joint number  and the optimal model size for efficient skeleton modeling. 
    \item We design a light-weight policy network and train it with the ST Gumbel Estimator to make it highly efficient. 
    \item We conduct extensive experiments on three benchmark datasets to demonstrate the superiority of our approach over state-of-the-art methods. Code will be released. 
\end{enumerate}

\section{Related Work}
\subsection{Skeleton-based action recognition}
Early-stage approaches for skeleton-based action recognition focus on designing various hand-crafted features~\cite{vemulapalli_human_2014}. 
With the success of the deep learning in the computer vision field, the data-driven methods have become the mainstream, which can be roughly divided into three kinds: 
RNN-based methods~\cite{zhang_view_2017,li_independently_2018,si_skeleton-based_2018,si_attention_2019}, CNN-based methods~\cite{li_skeleton-based_2017,liu_enhanced_2017,cao_skeleton-based_2018} and GCN-based methods~\cite{yan_spatial_2018,tang_deep_2018,shi_two-stream_2019,shi_skeleton-based_2019}. 
Recently, GCN-based methods have shown significant performance boost compared with other methods, indicating the necessity of the semantic human skeleton for action recognition. 
ST-GCN~\cite{yan_spatial_2018} is the first work to use GCN for skeleton-based action recognition, 
which aggregates the information of joints in a local area according to a predefined human body graph.
After that, Shi et al.~\cite{shi_two-stream_2019} propose an adaptive graph convolutional network to adapt the graph building process with the self-attention mechanism. They also introduce a two-stream framework to use both the joint information and the bone information.
Peng et al.~\cite{peng_learning_2020} turn to Neural Architecture Search (NAS) to automatically design GCN for skeleton-based action recognition.
Liu et al.~\cite{liu_disentangling_2020} disentangle the importance of nodes in different neighborhoods for effective long-range modeling. They also leverage the features of adjacent frames to capture complex spatial-temporal dependencies. 
However, these works did not consider an important factor for practical applications, i.e., the speed of the model. 
In contrast, our approach considers both the accuracy and the efficiency, which can adaptively adjust their trade-offs on demand. 

\subsection{Efficient Skeleton-based action recognition}
Recently, there are some works trying to design efficient models for skeleton-based action recognition~\cite{cheng_skeleton-based_2020,zhang_semantics-guided_2020}. 
Cheng et al.~\cite{cheng_skeleton-based_2020} (ShiftGCN) introduce two kinds of spatial shift graph operations to replace the heavy regular graph convolutions, which is more efficient and achieves strong performance. They also propose two kinds of temporal shift graph operation to outperform regular temporal models with less computational complexity. 
Zhang et al.~\cite{zhang_semantics-guided_2020} (SGN) propose to explicitly exploit the high-level information, i.e., the joint type and the frame index, to enhance the feature representation capability. 
Thus, they can use a smaller model to achieve a comparable performance compared with previous works. 
These methods mainly focus on designing light-weight modules. 
In contrast, our proposed method not only reduce the parameters of the model, but also reduce the amount of the data being processed, which is more efficient than previous works. 

\subsection{Efficient RGB-based Action Recognition}
For efficient RGB-based action recognition, there are mainly two streams of methods. 
One stream is to design light-weight or light-GFLOPs modules~\cite{qiu_learning_2017,xie_rethinking_2018,lin_tsm_2019,feichtenhofer_x3d_2020}. 
For example replacing the 3D convolution with the 2+1D convolutional operation~\cite{qiu_learning_2017} or replacing the temporal convolution with the temporal shift operation~\cite{lin_tsm_2019}. 
Another stream is the key-frame-based methods, which selects a small number of frames instead of all sequence for recognition to reduce the overall computational cost~\cite{zhu_key_2016,wu_adaframe_2019,korbar_scsampler_2019,wu_liteeval_2019,meng_ar-net_2020}. 
For example, Wu et al.~\cite{wu_adaframe_2019} propose to adaptively decide where to look next in an action sequence and whether the current prediction credible enough based on the current frame. They use recurrent models for policy network and use Reinforcement Learning methods to train it. 
Wu et al.~\cite{wu_multi-agent_2019} propose a multi-agent reinforcement learning framework, where each agent sample one frame based on the move instructions made by the policy network. 
Korbar et al.~\cite{korbar_scsampler_2019} introduce a lightweight ``clip-sampling" model to identify the most salient temporal clips for a long video. 
Meng et al.~\cite{meng_ar-net_2020} propose to adaptively select frames with the optimal resolution. They also use the recurrent models for policy network but train it with the Gumbel-Softmax tricks~\cite{jang_categorical_2017}, which has lower variance. 

While our approach is inspired by the key-frame-based methods, it is specially designed for the skeleton data, which can adaptively select not only the key skeletons (frames), but also the optimal model size and the optimal joint number for each human skeleton. 

\section{Methods}
\begin{figure*}[!htp]
    \centering
    \includegraphics[width=0.8\linewidth]{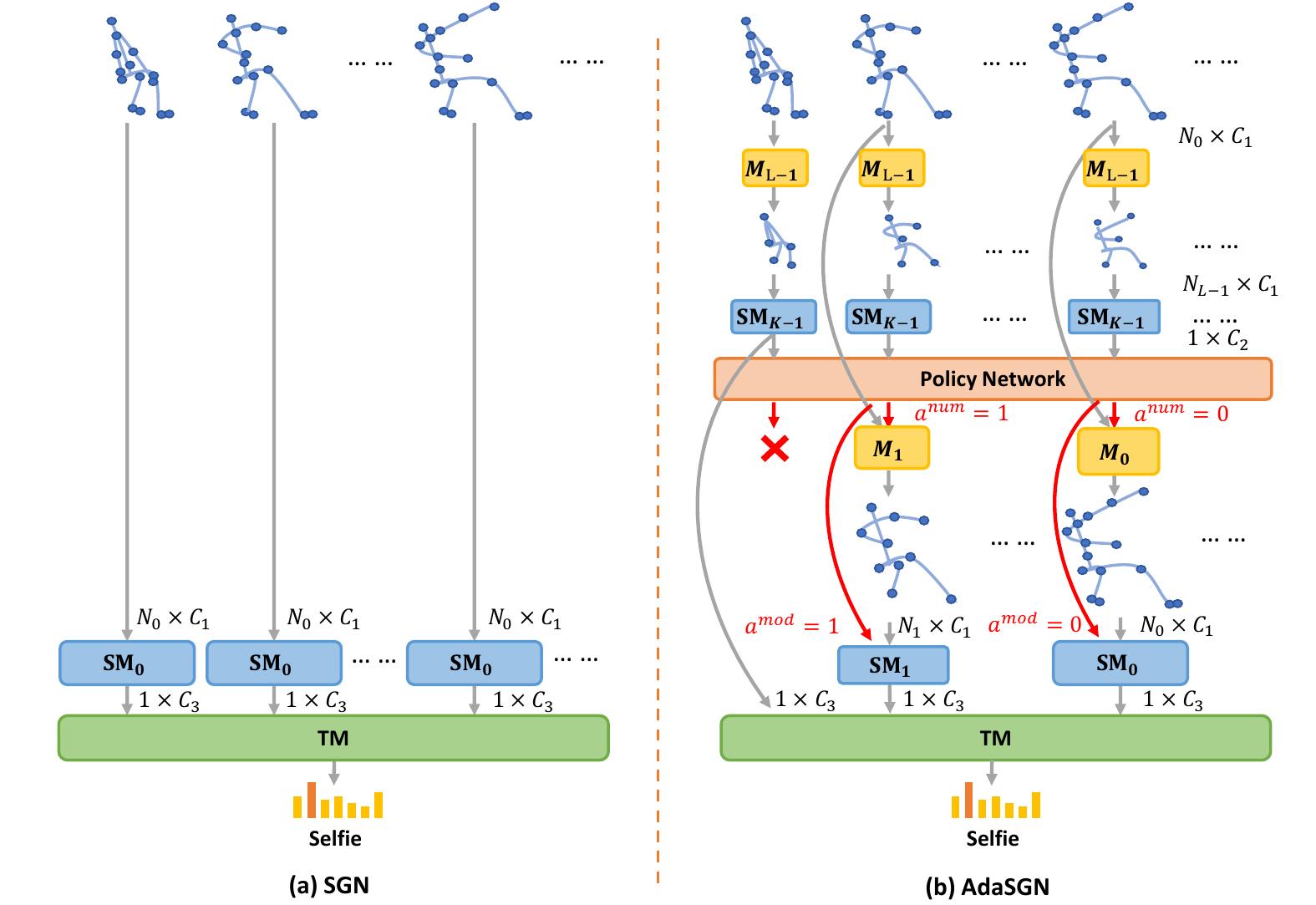}
    \caption{Pipeline of the SGN and the proposed AdaSGN. SM and TM denote the spatial module and the temporal module of the SGN. C denotes the channel number. 
    For AdaSGN, we assume that there are $K$ choices of joint number and $L$ choices of SM size. 
    Firstly, the raw input skeletons are aggregated into the skeletons with the smallest joint number ($N_{L-1}$) by transformation matrix $M_{L-1}$, which are fed into the smallest SM (i.e., SM$_{K-1}$) to extract policy features. 
    These policy features are fed into the policy network (orange box) to output actions ($a^{num}$ and $a^{mod}$) to decide the optimal joint number and SM size for each skeleton. 
    Then, each skeleton is transformed into the skeleton with the optimal joint number and modeled with the optimal-size SM to extract classification features. 
    Finally, the classification features of all skeletons are fed into the temporal network and the classifier to predict the final action label. }
    \label{fig:pipeline}
\end{figure*}

\subsection{Preliminaries}
\paragraph{Notations.}
A human action in skeleton format is represented as a skeleton sequence, which is denoted as 
$\mathcal{X}=\{ X_t\}_{t=0}^{T-1}$.
$X_t\in\mathbb{R}^{N\times C}$ is the 2D/3D coordinates of $N$ human key joints at time $t$. 
$T$ is the length of the sequence. 
$C$ denotes the coordinate dimension. 
If there are multiple persons, they are processed separately and the classification scores are averaged~\cite{yan_spatial_2018,shi_two-stream_2019}. 

\paragraph{GCN.}
Graph convolutional network (GCN) has been widely used for skeleton-based action recognition~\cite{yan_spatial_2018,shi_skeleton-based_2019}. 
A graph convolutional operation can be formulated as
\begin{equation}
\label{eq:gcn}
    Y_t=A_tX_tW
\end{equation}
where $A_t\in\mathbb{R}^{N\times N}$ is the normalized adjacency matrix, which denotes the graph topology. 
$W\in\mathbb{R}^{C_{in}\times C_{out}}$ denotes the convolutional weight. $X_t\in\mathbb{R}^{N\times C_{in}}$ and $Y_t\in\mathbb{R}^{N\times C_{out}}$ denote the input and output. 
One GCN layer consists of one graph convolutional operation, one BN operation~\cite{ioffe_batch_2015} and one ReLu activation function~\cite{nair_rectified_2010}. 
Multiple GCN layers can be stacked to enable further message passing among nodes with the same adjacency matrix.

In order to make the graph topology more flexible, $A_t$ can be adaptively learned to form the adaptive GCN layer~\cite{shi_skeleton-based_2019} as
\begin{equation}
\label{eq:attention}
    A_t = SoftMax(\theta(X_t)\phi(X_t)^T)
\end{equation}
where $\theta$ and $\phi$ denote two linear transformation functions, e.g., two 1-D convolutional layers whose kernel size is 1. 

\paragraph{SGN.}
Semantics-guided neural network (SGN)~\cite{zhang_semantics-guided_2020} is a light-weight GCN-based model for efficient skeleton-based action recognition, which is used as the baseline method in this paper. 
As shown in Figure~\ref{fig:pipeline} (left), SGN consists of multiple spatial modules (SM) and one temporal module (TM). 
SM (blue blocks) consists of three stacked adaptive GCN layers with the same adjacent matrix introduced in the last section, followed with a spatial max-pooling layer to aggregate joint features into one feature vector. 
The parameters of all SMs are shared. 
They are designed for exploiting the intra-frame correlations of joints. 
TM (green block) is designed for exploiting the correlations across the frames. 
It first concatenates the output features of SMs into one $T\times C$ matrix. 
Then two temporal convolutional layers (convolution along the temporal dimension) are appended to model the temporal dependency. 
After the TM, the feature maps are global-averaged and fed into a fully-connected layer for classification. 
More details that are irrelevant with our works such as the semantics embedding can refer to the original paper~\cite{zhang_semantics-guided_2020}.

\subsection{AdaSGN}
\paragraph{Pipeline overview.} Given a skeleton sequence, our goal is to select the optimal joint number and the optimal network size for each skeleton to efficiently extract features and predict actions. 
In this work, we use SGN~\cite{zhang_semantics-guided_2020} as the baseline model because it is already very efficient, which can better exhibit the effectiveness of the proposed method.  
For SGN, most of the computational costs come from SMs, so we modify the size of SM to control the computations of the whole model. 

Figure~\ref{fig:pipeline} (right) illustrates the overview of the proposed method called adaptive SGN (AdaSGN). 
Firstly, the raw input skeletons are transformed into the smallest-joint-number format, which are fed into the smallest SM to extract policy features. 
These policy features are fed into a light-weight policy network to output actions to decide the optimal joint number and the optimal SM size for every skeleton in the later steps. 
Note that this step brings nearly no additional computational cost compared with the later steps because of the light-weight modules and less joint number. 
Then, according to the decisions of the policy network, each skeleton is transformed into the skeleton with the optimal joint number and is modeled with the optimal-size SM to extract classification features. 
Finally, the classification features of all skeletons are fed into the temporal network to aggregate temporal correlations, whose outputs are used by the classifier to predict the final action label. 

\paragraph{Skeleton transformation.}
A skeleton sequence is denoted as $\mathcal{X}=\{X_t\in\mathbb{R}^{N\times C}\}$, where each frame consists of $N$ joints. 
AdaSGN adaptively chooses different joint numbers and model sizes to achieve efficiency. 
For different joint numbers, we define a sequence of joint numbers in descending orders as $\{N_i\}_{i=0}^{L-1}$. 
$N_0=N$ is the maximum number of joints. 
For different model sizes, we design $K$ SMs with different size in descending order, denoted as $\{$SM$_i\}_{i=0}^{K-1}$. 
SM$_0$ is the original SM in SGN. 

To transform the original skeleton with $N_0$ joints into skeletons with $N_i$ joints, we design a sequence of transformation matrices $\{M_i\in\mathbb{R}^{N_i\times N_0}\}$ as
\begin{equation}
    X_t^{N_i} = M_i X_t^{N_0}
    \label{eq:transform}
\end{equation}
where $X_t^{N_i}$ denotes skeleton $X_t$ with $N_i$ joints. 
The design principles of these transformation matrices are grouping semantics-adjacent joints (e.g., the hand and wrist) and average their coordinates. 
However, it is hard to ensure the optimality of the hand-crafted transformation matrix. 
Thus, we propose to set these transformation matrices as the model parameters and let the model to adaptively learn them. 
To stabilize the training in the beginning stage, we manually initialize these matrices and fixed them in the first several training epochs. 

\paragraph{Policy Network.}
Through the combination of different models and different joint numbers,  there are totally $K\times L$ choices with different GFLOPs for each skeleton, which formulates our action space. 
We first use the smallest number of joints, i.e., $N_{L-1}$, and the smallest SM, i.e., SM$_{K-1}$, to extract the policy features of each skeleton as
\begin{equation}
    F_t^{pol}=\textrm{SM}_{K-1}(M_{L-1}X_t^{N_0})
\end{equation}
The policy features of all frames are concatenated as a $T\times C$ matrix, which is fed into the policy network $\mathit{f}$ to output the action probabilities $P\in\mathbb{R}^{T\times (KL)}$ as
\begin{equation}
    P = SoftMax(\mathit{f}(Concat(\{F_t^{pol}\}_{t=0}^{T-1})))
\end{equation}
Here, we tried different modules for $\mathit{f}$, such as LSTM, Transformer and temporal convolution. 
The temporal convolution (convolution along the temporal dimension) is used in the final because it achieves the best performance. 

Given the probabilities, we can get the discrete actions $a$ through $\argmax$. 
However, directly perform $\argmax$ is not differentiable. 
Here, we use the Straight-Through (ST) Gumbel Estimator~\cite{jang_categorical_2017} to solve this problem. 
In detail, during the forward process, we use Gumbel-Max to sample a skeleton-level action according to the action probabilities
\begin{equation}
    a_t = \argmax_i(logP_{t, i}+G_{t, i})
\end{equation}
where $G_{t, i}=-log(-logU_{t, i})$ is a standard Gumbel distribution. $U_{t, i}$ is sampled from a uniform i.i.d distribution, i.e., $U_{t, i} \sim Uniform(0, 1)$. 
Because $\argmax$ is non-differentiable, in order to back-propagating gradients to the policy network, the continuous Gumbel-SoftMax is used to relax the Gumbel-Max in the back-propagation process as 
\begin{equation}
    \tilde{a_t} = \frac{\exp{(logP_{t, i} + G_{t, i})/\tau}}{\sum_{j=0}^{KL} \exp{(logP_{t, j} + G_{t, j})/\tau}}
\end{equation}
By denoting $\hat{a_t}=one\_hot(a_t)$ as the one-hot action vector, we treat $\tilde{a_t}$ as the continuous approximation of $\hat{a_t}$, i.e., $\nabla_\theta \tilde{a_t}\thickapprox \nabla_\theta \hat{a_t}$.
$\tau$ is the temperature parameter. 
For small temperatures, samples from the Gumbel-SoftMax are close to one-hot, i.e., it is more similar with the Gumbel-Max, but the variance of the gradients is large. 
For large temperatures, the gradients of samples from Gumbel-SoftMax have small variance, but it is more smooth and is more biased compared with the Gumbel-Max in the forward process. 
In practice, to balance the trade-off between the variance and bias, we initialize the $\tau$ to a high value and gradually anneal it down to a small value during the training as in \cite{jang_categorical_2017}. 

\paragraph{Classification}
After obtaining the action $a_t$ for each skeleton, 
we split it into the model action $a_t^{mod}$ and the joint number action $a_t^{num}$, where $a_t = a_t^{mod}\times L + a_t^{num}$.
Then we calculate the classification features of each skeleton by
\begin{equation}
    F_t^{cls}=\textrm{SM}_{a_t^{mod}}(M_{a_t^{num}}X_t^{N_0})
\end{equation}
If $a_t^{num}=L-1$ and $a_t^{mod}=K-1$, we directly use the policy features $F_t^{pol}$ instead of computing it again to reduce the computational cost (like the red fork in Figure~\ref{fig:pipeline}, right). 
The classification features are fed into the TM and the classifier $g$ to get the final output as 
\begin{equation}
    Y = SoftMax(\mathit{g}(\textrm{TM}(\{F_t^{cls}\}_{t=0}^{T-1})))
\end{equation}
where $g$ consists of one global-max-pooling layer and one fully-connected layer. 

\subsection{Loss Function}
The total loss function consists of two terms: one accuracy loss ($\mathcal{L}^{acc}$) and one efficiency loss ($\mathcal{L}^{eff}$) as
\begin{equation}
    \mathcal{L} = \mathcal{L}^{acc} + \alpha\mathcal{L}^{eff}
\end{equation}
where $\alpha$ is used to control the trade-off between the accuracy and the efficiency. 

$\mathcal{L}^{acc}$ is the standard cross-entropy loss as
\begin{equation}
    \mathcal{L}^{acc}=\mathbb{E}_{(\mathcal{X},y)}\sim\mathcal{D}^{train}(-ylog(\mathcal{F}(\mathcal{X};\Theta)))
\end{equation}
where $(\mathcal{X},y)$ is the training skeleton sequences and the associate one-hot action label. $\mathcal{F}$ denotes the model and $\Theta$ denotes the model parameters. 
$\mathcal{D}^{train}$ denotes the training data. 
$\mathcal{L}^{acc}$ only effect the classification quality of the model. 

$\mathcal{L}^{eff}$ controls the computational cost of the model. 
Because we use different sizes of networks and different number of joints based on the choices of the policy network, we calculate the GFLOPs of all these choices and use the mean GFLOPs as the loss term to encourage the less-computation operations as
\begin{equation}
    \mathcal{L}^{eff}=\mathbb{E}_{(\mathcal{X},y)}\sim\mathcal{D}^{train}(\frac{1}{T}\sum_{t=0}^{T-1}GFLOPS^\mathcal{F}(a_t))
\end{equation}
where $GFLOPs^\mathcal{F}$ is the pre-build GFLOPs lookup table for all action choices.

\section{Experiments}

\subsection{Dataset}

We perform extensive experiments on NTU-60~\cite{shahroudy_ntu_2016}, NTU-120~\cite{liu_ntu_2019} and SHREC~\cite{de_smedt_shrec17_2017}. NTU-60 recommends two benchmarks: cross-subject (CS) and cross-view (CV). NTU-120 recommends two benchmarks: cross-subject (CS) and cross-setup (CE). SHREC recommends two benchmarks: 14 gestures (14G) and 28 gestures (28G). Details of these datasets are provided in the supplementary material. 



\subsection{Implement Details}
Due to the space limitation, the complete training scheme and the architecture details are provided in the supplement material. 
We choose three ($L=3$) joint numbers ($N_i=\{1, 9, 25\}$ for NTU-60/120 and $N_i=\{1, 11, 22\}$ for SHREC) and two ($K=2$) model sizes (SM$_0$ and SM$_1$). 
Thus, there are totally six choices, i.e., $a_t=\{0, 1, 2, 3, 4, 5\}$, for policy network. 
When training, we first pretrain the single models with different number of joints. 
We show the  GFLOPs and accuracies of the six choices on NTU-60 on Table~\ref{tab:single_model}. 
* denotes adding the transform matrix as introduced in Eq.~\ref{eq:transform}. 
Compared with the original SGN, adaptively learning the transform matrix (SGN*) brings consistent improvements on both CS and CV benchmarks. 
It means it is better to appropriately adjust the original input skeletons. 
Besides, using less joints (9 or 1) reduce the GFLOPs of the model, but it also causes the drop of the accuracy. 

\begin{table}[!htb]
    \centering
    \caption{Recognition accuracy and GFLOPS of single models (\%) with different number of joints on NTU-60. * denotes adaptively transforming the number of joints into target numbers. }
    \label{tab:single_model}
    \vspace{0.2cm} 
    \begin{tabular}{l cc cc }
        \toprule\toprule
        Methods & \#joints & CS(\%) & CV(\%) & GFLOPs\\
        \midrule
        SM$_1$   & 25   & 87.1 & 92.9 & 0.078 \\
        SM$_1$*  & 25  & 87.9 & 93.3 & 0.078 \\
        SM$_1$*  & 9   & 86.7 & 92.3 & 0.033 \\
        SM$_1$*  & 1   & 63.6 & 74.1 & 0.009 \\
        \midrule
        SM$_0$   & 25   & 88.4 & 94.1 & 0.160 \\
        SM$_0$*  & 25  & 88.9 & 94.5 & 0.160 \\
        SM$_0$*  & 9   & 87.2 & 92.9 & 0.062 \\
        SM$_0$*  & 1   & 64.9 & 74.7 & 0.013 \\
        \bottomrule\bottomrule
    \end{tabular}
\end{table}

After pretraining the single models, we first load the pretrained parameters of the transform matrices and the SMs of these single models for AdaSGN. 
Then, it is trained using the same training schemes as the single models. 
For policy network, the $\tau$ of the Gumbel-SoftMax is initialized to 5 and linearly reduced ($\times 0.096$) at each epoch. 
$\alpha$ is gradually increased from 0 to the target value during the first 5 epochs. 
It is because we hope the model can focus more on the accuracy in the beginning training stage. 
Table~\ref{tab:scheme} shows the importance of the pretraining on NTU-60 dataset. 
We set $\alpha=4$ when not using pretrained models (i.e., ``w/o pretrain") for both CS and CV benchmarks. To keep the same GFLOPS for fair comparison, we set $\alpha=4$ and $\alpha=1$ when using pretrained models (i.e., ``w/ pretrain") for CS benchmark and CV benchmark, respectively. 
It shows using pretrained models can bring more than $2\%$ improvement of the accuracy with the same GFLOPs. 

\begin{table}[!htb]
    \centering
    \caption{Recognition accuracy (\%) and GFLOPS of Ada-Ske using different training scheme on NTU-60. ``w/o" and ``w/" denote ``without" and ``with".}
    \label{tab:scheme}
    \renewcommand\tabcolsep{4.0pt} 
    \vspace{0.2cm} 
    \begin{tabular}{l cc cc}
        \toprule\toprule
        \multirow{2}*[-3pt]{Strategy} & \multicolumn{2}{c}{CS} & \multicolumn{2}{c}{CV} \\
        \cmidrule{2-5}
        ~ & ACC(\%) & GFLOPS & ACC(\%) & GFLOPS\\
        \midrule
        w/o pretrain          & 87.0 & 0.07   & 92.2 & 0.10\\
        w/ pretrain      & 89.1 & 0.07   & 94.7 & 0.10\\
        \bottomrule\bottomrule
    \end{tabular}
\end{table}



\subsection{Trade-Off between ACC(\%) and Efficiency}

\begin{table*}[!htb]
    \centering
    \caption{Recognition accuracy (\%), GFLOPS and Percentages (\%) of the actions of Ada-Ske using different weights ($\alpha$) for efficient loss on NTU-60. ``S" and ``B" denotes S-SGCN and SM$_0$, repectively. 1, 9 and 25 is the number of joints. ``rand" denotes randomly select}
    \label{tab:trade_off}
    \vspace{0.2cm} 
    \renewcommand\tabcolsep{2.5pt} 
    \begin{tabular}{c cccccccc cccccccc}
        \toprule\toprule
        \multirow{3}{*}[-4pt]{$\alpha$} & \multicolumn{8}{c}{CS} & \multicolumn{8}{c}{CV} \\
        \cmidrule(lr){2-9} \cmidrule(lr){10-17}
        ~ & \multirow{2}{*}[-3pt]{ACC(\%)} & \multirow{2}{*}[-3pt]{GFLOPs} & \multicolumn{6}{c}{Actions(\%)} & \multirow{2}{*}[-3pt]{ACC(\%)} & \multirow{2}{*}[-3pt]{GFLOPs} & \multicolumn{6}{c}{Actions(\%)}
        \\
        \cmidrule(lr){4-9} \cmidrule(lr){12-17}
        ~ & ~ & ~ & S-1 & B-1 & S-9 & B-9 & S-25 & B-25 & ~ & ~ & S-1 & B-1 & S-9 & B-9 & S-25 & B-25 \\
        \midrule
        0.0 
        & 88.8 & 0.16 
        & 0.01 & 0.01 & 0.00 & 0.01 & 0.01 & 99.96
        & 94.6 & 0.16 
        & 0.01 & 0.01 & 0.01 & 0.01 &	0.01 & 99.95 \\
        0.1 
        & 88.9 & 0.12
        & 0.08 & 0.05 & 33.96 & 0.06 & 0.12 & 65.72
        & 94.7 & 0.16 
        & 0.02 & 0.01 &	0.01 & 0.01 & 0.01 & 99.94\\
        0.5 
        & 88.9 & 0.10 
        & 24.55& 0.12& 16.28& 0.09& 0.07 & 58.90
        & 94.8 & 0.10 
        & 40.68 & 0.01 & 0.01 & 0.01 & 0.01 & 59.29 \\
        1.0 
        & 88.9 & 0.10
        & 41.04 & 0.01 & 0.01 & 0.01 & 0.01 & 58.91
        & 94.7 & 0.10 
        & 42.75 & 0.03 & 0.68 & 0.03 & 0.02 & 56.49 \\
        2.0 
        & 89.0 & 0.09
        & 44.70 & 0.02 & 0.02 & 0.02 & 0.02 & 55.23
        & 94.7 & 0.09 
        & 44.18 & 0.05 & 0.01 & 7.77 & 0.00 & 47.98 \\
        4.0 
        & 89.1  & 0.07 
        & 48.36 & 0.01 & 0.01 & 16.47 & 0.01 & 35.14
        & 94.6  & 0.07 
        & 52.98 & 4.07 & 0.03 & 0.02 & 0.02 & 42.89 \\
        8.0 
        & 87.2  & 0.04
        & 52.76 & 0.01 & 0.01 & 47.21 & 0.01 & 0.00
        & 93.0  & 0.05 
        & 48.16 & 1.17 & 0.02 & 50.65 & 0.00 & 0.01 \\
        10.0 
        & 86.9  & 0.03 
        & 56.73 & 0.01 & 0.01 & 43.25 & 0.01 & 0.00
        & 92.8  & 0.04 
        & 52.09 & 1.40 & 0.01 & 46.49 & 0.01 & 0.00 \\
        20.0 
        & 85.9 & 0.02 
        & 73.27 & 0.03 & 0.00 & 26.68 & 0.00 & 0.00 
        & 91.6 & 0.03 
        & 77.40 & 3.41 & 0.00 &	14.45 &	0.00 & 4.74 \\
        \midrule
        rand
        & 85.6 & 0.06 
        & 16.7 & 16.7 & 16.7 & 16.7 & 16.7 & 16.7 
        & 91.2 & 0.06 
        & 16.7 & 16.7 & 16.7 & 16.7 & 16.7 & 16.7  \\
        fuse 
        & 89.3 & 0.32
        & 100 & 100 & 100 & 100 & 100 & 100 
        & 94.9 & 0.32 
        & 100 & 100 & 100 & 100 & 100 & 100  \\
        \bottomrule\bottomrule
    \end{tabular}
\end{table*}

By adjusting the $\alpha$, we can control the trade-off between accuracy and efficiency. 
We show the accuracy, GFLOPS and the percent of action choices for different $\alpha$ on Table~\ref{tab:trade_off}. 
$B$ and $S$ denote using the big SM ($SM_0$) and the small SM ($SM_1$), respectively. 
With the increase of $\alpha$, the GFLOPS decreases, and the accuracy first increases and then decreases. 
It is because most of the joints in the skeleton data are redundant, appropriately reducing the joint points can reduce the noise and make the model pay more attention to the important features. 
When $\alpha=0$, only the loss of accuracy works, so AdaSGN uses only the biggest model ($SM_0$) and all of the joints (25).
When increasing $\alpha$, it chooses different number of joints and different models for different samples and frames to balance the accuracy and efficiency.
For CS benchmark, when set $\alpha=4$, it achieves a good trade-off between the accuracy and efficiency. 
It achieves $0.3\%$ improvement of accuracy with only $43.8\%$ GFLOPS compared with $\alpha=0$.
For CV benchmark, when set $\alpha=0.5$, it obtains $0.2\%$ improvement of accuracy with only $62.5\%$ GFLOPS compared with $\alpha=0$. 

Besides, we test two other policies, i.e., ``rand" and ``fuse". 
``rand" denotes randomly selecting the actions, where the probability of every action choice is 16.7\%. 
It achieves lowest performance with medium GFLOPs. Compared with $\alpha=4$ setting, it needs nearly the same GFLOPs (0.06 vs. 0.07), but it achieves much lower accuracies ($85.6\%$ vs. $89.1\%$ on CS and $91.2\%$ vs. $94.6\%$ on CV). 
``fuse" denotes directly fusing the scores of all the models, where the probability of every action choice is 100\%. It needs much more GFLOPs because it calculates all models. 
Compared with $\alpha=4$, it brings only $0.2\%/0.3\%$ improvements on CS/CV benchmark, but it requires $4.6$ times GFLOPs. 
These two comparisons confirm the necessary of adaptively selecting suitable actions for different samples and frames. 

\begin{figure}
    \centering
    \includegraphics[width=\linewidth]{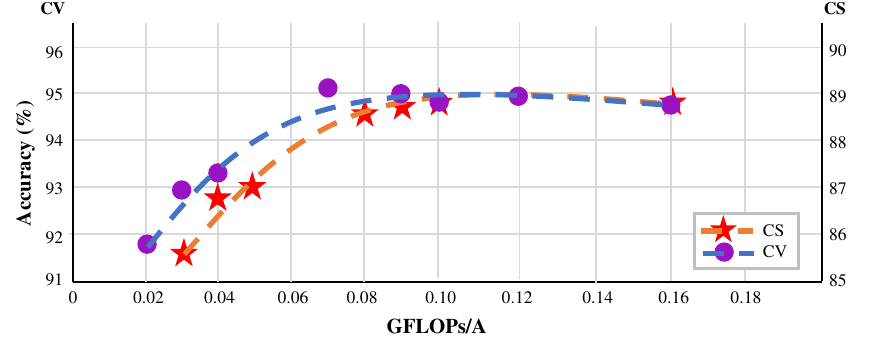}
    \caption{GFLOPs v.s. accuracy for AdaSGN on NTU-60 dataset. }
    \label{fig:flopntu}
\end{figure}

For better visualization, we plot the accuracy-efficiency trade-off of NTU-60 on Figure~\ref{fig:flopntu}. Using this curve we can better set a good $\alpha$ depending on the demand. 
We also plot similar curves for NTU-120 and SHREC in the supplement material, which shows consistent results. 

\begin{figure*}[!htp]
    \centering
    \includegraphics[width=0.9\linewidth]{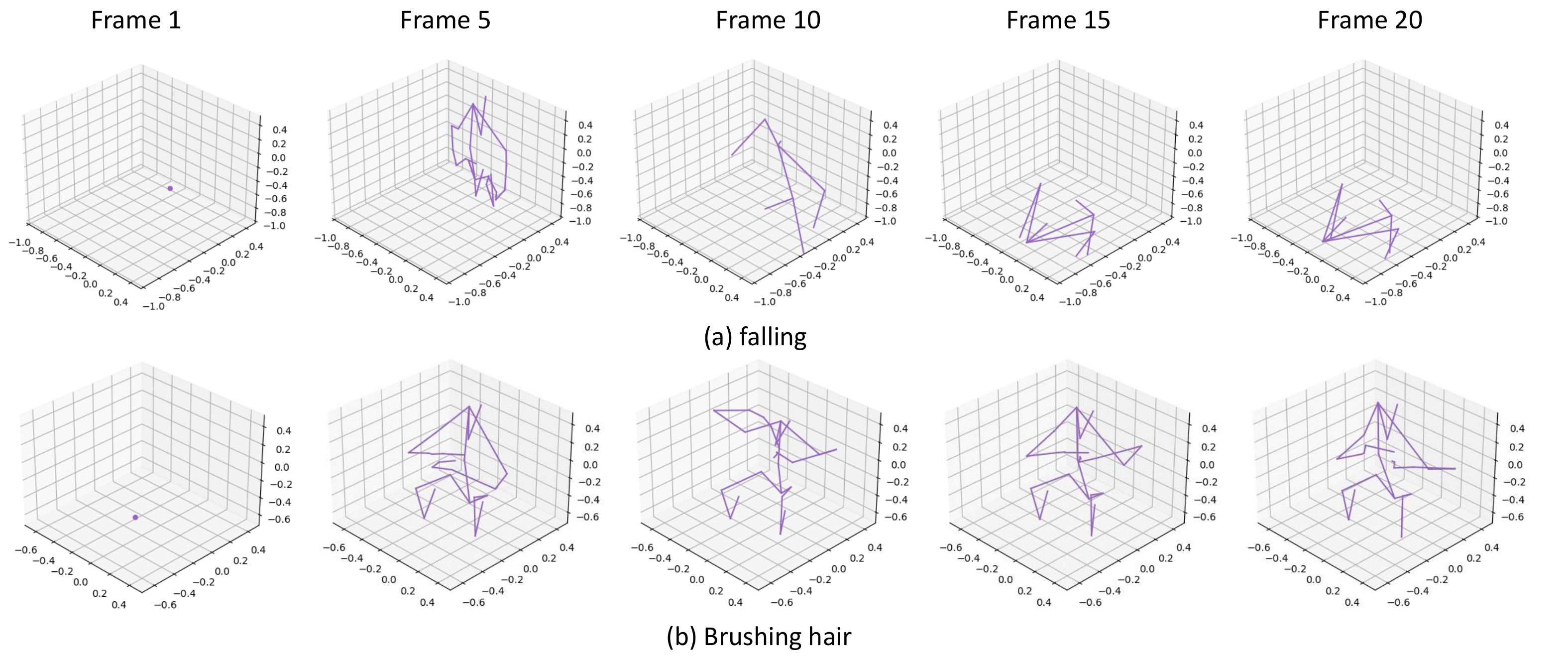}
    \caption{Qualitative examples from NTU-60. Each sample has 20 frames and we show 5 of them evenly. Non-informative skeletons are transformed into skeletons with less points while other informative skeletons are kept with the original point numbers. }
    \label{fig:example}
\end{figure*}

\subsection{Comparison with SOTA}

\begin{table}[!htb]
    \centering
    \caption{Comparison with the state-of-the-arts on NTU-60.}
    \label{tab:NTU}
    \vspace{0.2cm} 
    \renewcommand\tabcolsep{3.0pt} 
    \begin{tabular}{l cc cc}
        \toprule\toprule
        \multirow{2}*[-3pt]{Methods} & \multicolumn{2}{c}{CS} & \multicolumn{2}{c}{CV} \\
        \cmidrule{2-5}
        ~ & ACC(\%) & GFLOPS & ACC(\%) & GFLOPS\\
        \midrule
        ST-GCN~\cite{yan_spatial_2018} & 81.5 & 16.3 &  88.3 & 16.3 \\
        ASGCN~\cite{li_actional-structural_2019} & 86.8 & 27.0 & 94.2 & 27.0\\
        AGCN~\cite{shi_two-stream_2019} & 88.5 & 35.8 & 95.1 & 35.8\\
        MS-G3D~\cite{liu_disentangling_2020} & 91.5 & 48.8 & 96.2 & 48.8 \\
        DSTA-Net~\cite{shi_decoupled_2020} & 91.5 & 55.6 & 96.4 & 55.6 \\
        \midrule
        ShiftGCN~\cite{cheng_skeleton-based_2020}        & 90.7 & 10.0   & 96.5 & 10.0\\
        SGN~\cite{zhang_semantics-guided_2020}        & 89.0 & 0.80   & 94.5 & 0.80\\
        \midrule
        1s-SGN~\cite{zhang_semantics-guided_2020}        & 88.4 & 0.16   & 94.1 & 0.16\\
        \textbf{1s-AdaSGN}        & 89.1 & 0.07   & 94.6 & 0.07 \\
        \midrule
        3s-SGN~\cite{zhang_semantics-guided_2020}        & 90.0 & 0.48   & 94.9 & 0.48\\
        \textbf{3s-AdaSGN}       & 90.5 & 0.19   & 95.3 & 0.15 \\
        \bottomrule\bottomrule
    \end{tabular}
\end{table}
We compare the proposed AdaSGN with the other state-of-the-art approaches on NTU-60/120 and SHREC datasets as shown in Table~\ref{tab:NTU} and Table~\ref{tab:shrec}. 
Original SGN randomly select 5 sequences and average the scores to obtain the prediction. It is denoted as ``SGN". 
We implement it using one sequence for fair comparison and denote it as ``1s-SGN". 
Because the ``bone" and ``velocity" information has been shown great complementarity with the original data~\cite{shi_two-stream_2019,shi_decoupled_2020}, we fuse them in a three-stream framework, which is denoted as``3s-SGN".
In detail, we train three models using the ``joint", ``bone" and ``velocity" respectively and average their classification scores for prediction. 
``1s-AdaSGN" and ``3s-AdaSGN" are named in the same manner. 

Results of NTU-60 are shown in Table~\ref{tab:NTU}. 
The first group of approaches, i.e., ST-GCN~\cite{yan_spatial_2018}, ASGCN~\cite{li_actional-structural_2019}, etc, only consider the performance; thus, they use a large amount of computational budgets. 
The second group of approaches, i.e., ShiftGCN~\cite{cheng_skeleton-based_2020} and SGN~\cite{zhang_semantics-guided_2020}, are designed for both accuracy and efficiency. They achieve slightly lower accuracy but much faster speed. 
Our methods (AdaSGN) are based on SGN.
Compared with 1s-SGN, 1s-AdaSGN ($\alpha=4$) achieves higher accuracy ($+0.7\%/+0.5\%$) with less than half of the GFLOPS (43.8\%/43.8\%) on CS/CV benchmarks. 
Similarly, 3s-AdaSGN also brings $+0.5\%/+0.4\%$ improvements with nearly a third of the GFLOPS (39.6\%/31.3\%) on CS/CV benchmarks.
Compared with the DSTA-Net~\cite{shi_decoupled_2020} which achieves highest accuracy, our method is $200$ times faster with only $1\%$ drop of accuracy. 
We also plot the scatter diagram on Figure~\ref{fig:scatter}, which intuitively shows the superiority of our method. 

\begin{table}[!htb]
    \centering
    \caption{Comparison with the state-of-the-arts on NTU-120.}
    \label{tab:NTU120}
    \vspace{0.2cm} 
    \renewcommand\tabcolsep{3.0pt} 
    \begin{tabular}{l cc cc}
        \toprule\toprule
        \multirow{2}*[-3pt]{Methods} & \multicolumn{2}{c}{CS} & \multicolumn{2}{c}{CE} \\
        \cmidrule{2-5}
        ~ & ACC(\%) & GFLOPS & ACC(\%) & GFLOPS\\
        \midrule
        AGCN~\cite{shi_two-stream_2019} & 82.9 & 35.8 & 84.9 & 35.8\\
        MS-G3D~\cite{liu_disentangling_2020} & 86.9 & 48.8 & 88.4 & 48.8 \\
        DSTA-Net~\cite{shi_decoupled_2020} & 86.6 & 55.6 & 89.0 & 55.6 \\
        \midrule
        ShiftGCN~\cite{cheng_skeleton-based_2020}        & 85.9 & 10.0   & 87.6 & 10.0\\
        SGN~\cite{zhang_semantics-guided_2020}        & 79.2 & 0.80   & 81.5 & 0.80\\
        \midrule
        1s-SGN~\cite{zhang_semantics-guided_2020}        & 82.1 & 0.16   & 82.2 & 0.16\\
        \textbf{1s-AdaSGN}        & 83.3 & 0.08   & 83.6 & 0.08 \\
        \midrule
        3s-SGN~\cite{zhang_semantics-guided_2020}       & 85.5 & 0.48   & 86.3 & 0.48\\
        \textbf{3s-AdaSGN}       & 85.9 & 0.21   & 86.8 & 0.26 \\
        \bottomrule\bottomrule
    \end{tabular}
\end{table}

\begin{table}[!htb]
    \centering
    \caption{Comparison with the state-of-the-arts on SHREC.}
    \label{tab:shrec}
    \vspace{0.2cm} 
    \renewcommand\tabcolsep{3.0pt} 
    \begin{tabular}{l cc cc}
        \toprule\toprule
        \multirow{2}*[-3pt]{Methods} & \multicolumn{2}{c}{14G} & \multicolumn{2}{c}{28G} \\
        \cmidrule{2-5}
        ~ & ACC(\%) & GFLOPS & ACC(\%) & GFLOPS\\
        \midrule
        ST-GCN~\cite{yan_spatial_2018} & 92.7 & 7.2 &  87.7 & 7.2 \\
        HPEV~\cite{liu_decoupled_2020} & 94.9 & 1.46 & 92.3 & 1.46 \\
        DSTA-Net~\cite{shi_decoupled_2020} & 97.0 & 14.4 & 93.9 & 14.4 \\
        \midrule
        1s-SGN~\cite{zhang_semantics-guided_2020}        & 94.8 & 0.15   & 92.3 & 0.15\\
        \textbf{1s-AdaSGN}        & 94.9 & 0.05   & 92.3 & 0.05 \\
        \midrule
        3s-SGN~\cite{zhang_semantics-guided_2020}        & 96.3 & 0.45   & 93.8 & 0.45\\
        \textbf{3s-AdaSGN}        & 96.3 & 0.21   & 94.0 & 0.23 \\
        \bottomrule\bottomrule
    \end{tabular}
\end{table}

We also compared our method with the state-of-the-arts on NTU-120, which is larger and more challenging compared with NTU-60, and SHREC, which is used for skeleton-based hand gesture recognition. 
As shown in Table~\ref{tab:NTU120} and Table~\ref{tab:shrec}, these two datasets show consistent results with NTU-60, which further verified the effectiveness and generalizability of our method. 

\subsection{Qualitative Results}

Figure~\ref{fig:example} shows some qualitative examples from NTU-60 dataset. 
We uniformly sample 5 skeletons from each skeleton sequence and show them in the 3-dimension coordinate system. Note that the coordinates of the human joints are transformed by the adaptively learned transformation matrix. 
They are a little different with the original joints in the camera coordinate system. 

It shows the policy network choose different number of joints for different samples and different skeletons. 
For example, the first skeleton of the skeleton sequence in NTU is always the action independent preparation stage; thus, the policy network skip these skeletons. 
For action ``falling", the action is mainly determined by the posture of the whole body. 
So the policy network uses a part of joints (9-joint) for later skeletons. 
But for action ``brushing hair", the action is recognized based on the details of the hand and head. Thus, the policy network keeps the original number of joints (25-joint) for later skeletons. 
More qualitative results are provided in the supplement material. 

\section{Conclusion}
In this paper, we propose a novel approach, called AdaSGN, for efficient skeleton-based action recognition.
It can adaptively select the optimal joint number and the optimal model size for each skeleton to balance the trade-off between the accuracy and the efficiency. 
A light-weight policy network is designed to generate such selections and is trained with the ST Gumbel Estimator to make this process highly efficient. 
To verify the superiority of the proposed methods, extensive experiments are performed on three challenging datasets, where our method achieves consistent improvement in accuracy and significant savings in GFLOPs. 

Future works can explore how to adaptively select the specific joint instead of the number of joints, which should be more flexible for various samples. 
Besides, it is also worth to exploring how to achieve the stepless adjustment for both the model size and the joint numbers, which can avoid the manual settings of the candidates. 
{\small
\bibliographystyle{ieee_fullname}
\bibliography{references}
}

\end{document}